\documentclass[sigconf]{acmart}

\AtBeginDocument{%
  \providecommand\BibTeX{{%
    \normalfont B\kern-0.5em{\scshape i\kern-0.25em b}\kern-0.8em\TeX}}}

\copyrightyear{2022}
\acmYear{2022}
\setcopyright{acmcopyright}\acmConference[KDD '22]{Proceedings of the 28th ACM SIGKDD Conference on Knowledge Discovery and Data Mining}{August 14--18, 2022}{Washington, DC, USA}
\acmBooktitle{Proceedings of the 28th ACM SIGKDD Conference on Knowledge Discovery and Data Mining (KDD '22), August 14--18, 2022, Washington, DC, USA}
\acmPrice{15.00}
\acmDOI{10.1145/3534678.3539194}
\acmISBN{978-1-4503-9385-0/22/08}

\usepackage{booktabs}       
\newcommand{\mr}[2]{\multirow{#1}{*}{#2}}
\newcommand{\mc}[3]{\multicolumn{#1}{#2}{#3}}
\usepackage{amsfonts}       
\usepackage{nicefrac}       
\usepackage{microtype}      
\usepackage{array}
\newcolumntype{P}[1]{>{\centering\arraybackslash}p{#1}}
\usepackage{url}            

\usepackage{amsmath,amsthm,bm,amsmath} 
\usepackage{xspace}
\usepackage{enumerate}
\usepackage{enumitem}
\usepackage{stmaryrd}
\usepackage{caption}
\usepackage{subcaption}
\usepackage{multirow}
\usepackage{times}
\usepackage{graphicx} 
\usepackage{sidecap}
\usepackage{wrapfig}
\usepackage{colortbl}
\usepackage{arydshln}
\usepackage{tikz}
\usepackage{listings}
\usepackage{xcolor}
\usepackage{makecell}
\usepackage{booktabs}
\usepackage{titletoc}
\usepackage{hyperref}
\usepackage{arydshln}
\usepackage{bm}

\newcommand{\method}{BC-Aligner}

\definecolor{customgray}{rgb}{0.3,0.3,0.3}
\definecolor{customgreen}{RGB}{140,211,89}

\newcommand{\vb}[1]{\textbf{\textit{#1}}}

\usepackage{hyperref}

\newcount\COMMENTs  
\usepackage{color}
\definecolor{darkgreen}{rgb}{0,0.5,0}
\definecolor{purple}{rgb}{1,0,1}
\newcommand{\comm}[2]{\ifnum\COMMENTs=1\textcolor{#1}{#2}\fi}

\newcommand{\xhdr}[1]{\vspace{0.1cm}{\noindent\bfseries #1}.}

\definecolor{dkred}{rgb}{0.5,0,0}
\definecolor{dkgreen}{rgb}{0,0.6,0}
\definecolor{gray}{rgb}{0.5,0.5,0.5}
\definecolor{mauve}{rgb}{0.58,0,0.82}

\newcommand{\ie}{\textit{i.e.}}
\newcommand{\eg}{\textit{e.g.}}

\begin{document}

\title{Learning Backward Compatible Embeddings}







\author{Weihua Hu$^\dagger$, Rajas Bansal$^\dagger$, Kaidi Cao$^\dagger$, Nikhil Rao$^\ddagger$, Karthik Subbian$^\ddagger$, Jure Leskovec$^\dagger$}
\affiliation{%
  \institution{$^\dagger$Stanford University, $^\ddagger$Amazon}
  \country{}}
\email{{weihuahu,rajasb,kaidicao,jure}@cs.stanford.edu,  {nikhilsr,ksubbian}@amazon.com}

\renewcommand{\shortauthors}{Hu et al.}

\begin{abstract}
Embeddings, low-dimensional vector representation of objects, are fundamental in building modern machine learning systems. In industrial settings, there is usually an embedding team that trains an embedding model to solve {\em intended tasks} (\eg, product recommendation). The produced embeddings are then widely consumed by consumer teams to solve their {\em unintended tasks} (\eg, fraud detection). However, as the embedding model gets updated and retrained to improve performance on the intended task, the newly-generated embeddings are no longer compatible with the existing consumer models. 
This means that historical versions of the embeddings can never be retired or all consumer teams have to retrain their models to make them compatible with the latest version of the embeddings, both of which are extremely costly in practice.

Here we study the problem of embedding version updates and their backward compatibility. 
We formalize the problem where the goal is for the embedding team to keep updating the embedding version, while the consumer teams do not have to retrain their models.
We develop a solution based on learning backward compatible embeddings, which allows the embedding model version to be updated frequently, while also allowing the latest version of the embedding to be quickly transformed into any backward compatible historical version of it, so that consumer teams do not have to retrain their models.
Our key idea is that whenever a new embedding model is trained, we learn it together with a light-weight backward compatibility transformation that aligns the new embedding to the previous version of it. 
Our learned backward transformations can then be composed to produce any historical version of embedding.
Under our framework, we explore six methods and systematically evaluate them on a real-world recommender system application.
We show that the best method, which we call \method, maintains backward compatibility with existing unintended tasks even after multiple model version updates. Simultaneously, \method{} achieves the intended task performance similar to the embedding model that is solely optimized for the intended task.\footnote{Code is publicly available at \url{https://github.com/snap-stanford/bc-emb}}

\end{abstract}

\maketitle

\section{Introduction}
\label{sec:intro}
\begin{figure}[t]
  \centering
  \includegraphics[width=0.95\linewidth]{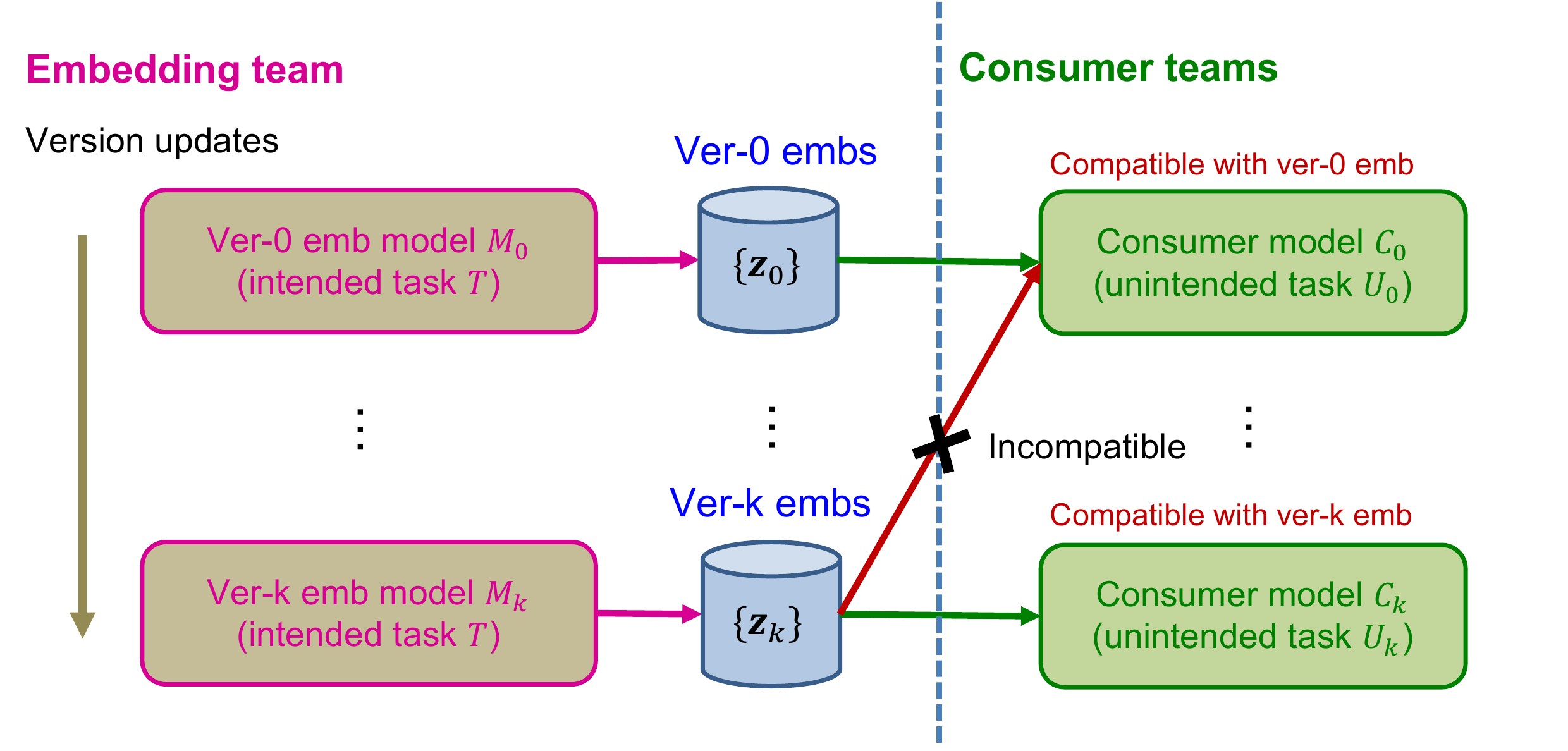}
  \caption{Problem formulation. The embedding team trains {\em embedding model} $M_0$ to solve their {\em intended task} $T$. The consumer teams may then utilize the produced embeddings $\{\vb{z}_0\}$ to solve some {\em unintended task} $U_0$ using consumer model $C_0$. The issue arises when the embedding team releases new improved versions of the embedding model $M_1, M_2, \ldots$ over time. At version $k$, the latest-version embedding model $M_k$ produces ver-$k$ embeddings $\{\vb{z}_k\}$ that are incompatible with consumer model $C_0$ that is trained on the ver-0 embeddings $\{\vb{z}_0\}$. Our goal is to quickly transform the latest-version embedding $\vb{z}_k$ into a backward compatible historical version of it so that existing consumer models can readily use the transformed embedding without being retrained.
  }
  \label{fig:task_framework}
    \vspace{-5mm}
\end{figure}

Embeddings are widely used to build modern machine learning systems. In the context of  recommender systems, embeddings are used to represent items and to capture similarity between them. Such embeddings can be then used for many tasks like item recommendations, item relevance prediction, item property prediction, item sales volume prediction as well as fraud detection~\citep{ying2018graph, wang2021bipartite}.

The universality of embeddings and the proliferation of state of the art methods to generate these embeddings~\citep{hu2020open} pose an interesting challenge. Many machine learning practitioners develop embeddings for a specific purpose (\eg, for item recommendation) but then the embeddings get utilized by many other downstream consumer teams for their own purposes and tasks. Oftentimes, the number of such consumer teams is very large and hard to track. At the same time, the original embedding team aims to further evolve their embedding model architecture, training data, and training protocols, with the goal to improve performance on their specific task. In this process, the embedding team generates new and improved versions of the embeddings but these are incompatible with existing downstream consumer models. This means the downstream consumers of these embeddings must retrain their models to use the improved embeddings, or choose to stick with the older, potentially poorer embeddings as inputs to their models. To maintain compatibility with all the existing consumer tasks, the embedding team needs to maintain all historical versions of their embedding model, generate all historical versions of the embeddings and make them available to the consumer teams. This means that the historical embedding models can never be retired and significant human and computational resources are wasted. An alternative approach would be to retire old versions of the embeddings and encourage the consumer teams to retrain their models and migrate to the new version of the embeddings. In practice, this is extremely hard. It can take years before all the consumer teams move to a new version of the embeddings, and the old versions can be retired. 
In general, the problem of backward incompatible embeddings slows down iteration cycles and leads to significant human and computing cost.

\xhdr{Present work: Backward compatible embeddings} Here we study the problem of evolving embedding models and their backward compatibility (Figure ~\ref{fig:task_framework}). We formalize a realistic setting where the embedding team works on developing an {\em embedding model} $M$ that is trained to predict a given {\em intended task} $T$ (intended for the embedding team), \eg, item recommendation. Over time, the embedding team adds new training data, keeps experimenting with different model architectures, embedding dimensions, and hyperparameters, which results in evolving \emph{versions} of the embedding model, $M_0, M_1, M_2, \ldots$, where we use the subscript to indicate the version.
Given an input data point $x$, each such embedding model $M_i$ produces its own $D_i$-dimensional embedding $\vb{z}_i \equiv M_i(x) \in \mathbb{R}^{D_i}$.
Notice that at version $k$, we have $k+1$ versions of embeddings $\vb{z}_0, \vb{z}_1, \ldots, \vb{z}_k$ for the same input data point $x$.
We call $\vb{z}_i$ \emph{ver-$i$ embedding} of $x$.
In practice, we have a collection of input data points, for which we use the embedding model to generate the embeddings. Such embeddings are then stored and shared with other teams.
We use $\{\}$ to denote the collection of the generated embeddings, \eg, $\{\vb{z}_k\}$ denotes the collection of ver-$k$ embeddings.
Note that $\{\vb{z}_k\}$ can be refreshed frequently as new data points arrive, \eg, item embeddings may be refreshed every day as new items/iteractions arrive.

\begin{figure}[t]
  \centering
  \includegraphics[width=0.9\linewidth]{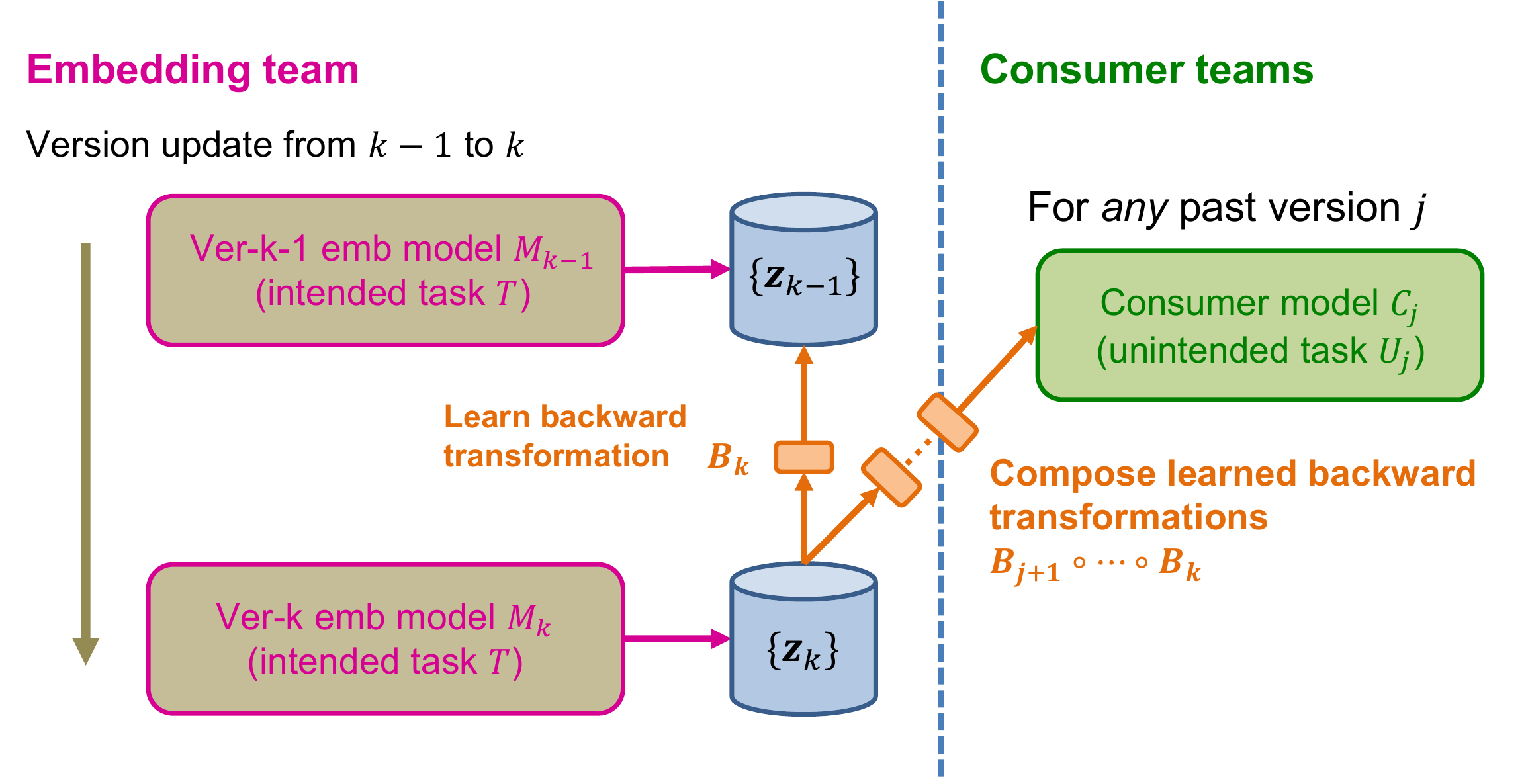}
  \vspace{-2mm}
  \caption{
  An overview of our framework. We train a new embedding model $M_k$ and a light-weight backward transformation function $B_k$ by optimizing the two training objectives simultaneously: (1) to solve the intended task $T$, and (2) to align ver-$k$ embeddings $\{\vb{z}_k\}$ to ver-$k-1$ embeddings $\{\vb{z}_{k-1}\}$ using $B_k$. We use the latest-version embedding model $M_k$ to produce embeddings $\{\vb{z}_k\}$ and store them.
  For any existing consumer model $C_j$ requesting a ver-$j$ compatible embedding $\widetilde{\vb{z}}_j$, we compose the learned backward transformations as $B_{j+1}\circ \cdots \circ B_k$ \emph{on-the-fly}, \ie, $\widetilde{\vb{z}}_j = B_{j+1}\circ \cdots \circ B_k(\vb{z}_k)$.
  }
  \vspace{-5mm}
  \label{fig:method_framework}
\end{figure}

At the same time, we have many other {\em consumer teams} that utilize the produced embeddings to solve their \emph{unintended tasks} (unintended for the embedding team), \eg, fraud detection.
Consider a consumer team that started to use the embeddings at some version $j$. They would use ver-$j$ embeddings to train their consumer model $C_j$ for their own unintended task $U_j$. However,  later ver-$k$ embeddings ($k > j$) are generally incompatible with consumer model $C_j$; hence, $C_j$ cannot simply use ver-$k$ embeddings as input. So, either the consumer model $C_j$ need to be retrained and recalibrated on the later ver-$k$ embeddings, or both the old ver-$j$ embeddings have to be also maintained. Both solutions lead to significant cost and overhead.

We first formalize the problem where the goal is for the embedding team to keep improving and updating the embedding model, while the existing consumer teams do not have to retrain their models when a new version of the embedding model is released. We then develop a solution based on learning backward compatible embeddings (Figure~\ref{fig:method_framework}), which allows the embedding model version to be updated frequently, while also allowing the latest-version embedding to be quickly transformed into any backward compatible historical version of it, so that the existing consumer teams do not have to retrain their models. Our key idea is that whenever a new embedding model $M_k$ is trained, we learn it together with a light-weight {\em backward (compatibility) transformation} $B_k: \mathbb{R}^{D_k} \to \mathbb{R}^{D_{k-1}}$ that aligns ver-$k$ embeddings $\vb{z}_k \in \mathbb{R}^k$ to its previous version $\vb{z}_{k-1}  \in \mathbb{R}^{k-1}$, \ie, $B_i(\vb{z}_{k}) \approx \vb{z}_{k-1}$. 
At version $k$, we maintain the learned $B_k$ as well as all past backward transformation functions learned so far: $B_1, B_2, \ldots, B_{k-1}$.

Importantly, our learned backward transformations can be composed to approximate any historical version of the embedding. Specifically, at version $k$, the latest-version embedding $\vb{z}_k$ can be transformed to approximate any historical ver-$j$ embedding $\vb{z}_j$ ($j<k$) by $B_{j+1} \circ \cdots \circ B_{k-1} \circ B_{k}(\vb{z}_k) \approx \vb{z}_j$, where $\circ$ denotes the function composition.
We call the transformed embedding \emph{ver-$j$ compatible embedding}, and denote it as $\widetilde{\vb{z}}_j$. Because $\widetilde{\vb{z}}_j \approx \vb{z}_j$, the embedding $\widetilde{\vb{z}}_j$ can be used by consumer model $C_j$ in a compatible manner.
The backward transformations are fast and lightweight so they can be applied on-the-fly whenever a consumer team requests a historical version of the latest-version embedding.
Furthermore our solution is fully inductive: given an unseen input data point $x$, the learned backward transformations can be used to quickly transform the newly-generated ver-$k$ embedding $\vb{z}_k \equiv M_k(x)$ into its historically compatible version $\widetilde{\vb{z}_j}$ for any $j < k$.

At version $k$, we use the linear model for $B_k$ and jointly train it with $M_k$ to solve the intended task $T$ as well as to encourage $B_k$ to align the embedding output of $M_k$ to that of $M_{k-1}$, \ie, $B_k \circ M_k \approx M_{k-1}$.
Additionally, we develop a novel loss that suppresses the amplification of the alignment error caused by the composition of backward transformations. Altogether, we arrive at our proposed method, which we call \method{}.
In addition, we consider five other method variants under our framework, with different design choices of $B_k$, loss function, and training strategy of $B_k$ and $M_k$, some of which includes prior methods~\citep{shen2020towards,mikolov2013distributed}.

To systematically evaluate different methods, we introduce a realistic experimental setting in the real-world recommender system application~\citep{ying2018graph}. We consider link prediction as the intended task $T$, graph neural networks as the embedding model $M$, and propose five different unintended tasks $U$ that are of practical interest.
We empirically show that \method{} provides the best performance compared to the other method variants including the prior works.
\method{} maintains nearly-perfect backward compatibility with existing unintended tasks even after multiple rounds of embedding model updates over time.  
Simultaneously, \method{} achieves the intended task performance comparable to the embedding model that is solely optimized for the intended task.

Overall, our work presents the first step towards solving the critical problem of incompatibility between the embedding model and unintended downstream consumer models.
We hope our work spurs an interest in community to solve the new problem setting.

\section{Problem Setting}
\label{sec:problem}
Here we formalize our problem setting.
We ground our application to recommender systems, though our formulation is general and applicable to broader application domains that use embeddings to perform a wide range of tasks, such as computer vision, natural language processing, search and information retrieval. The new concepts and terminology introduced in this paper are bolded.

\subsection{Terminology and Setting}
We consider two types of machine learning tasks: The \textbf{\emph{intended task}} $T$ and \textbf{\emph{unintended tasks}} $U$.
The intended task is the task that the embedding team originally intended to solve and is solved by using embeddings produced by a deep learning model, which we call the \textbf{\emph{embedding model}} $M$.
This embedding model is trained to solve the intended task $T$.
At the same time, these embeddings can be used by many consumer teams to solve their tasks that may not be originally intended by the embedding team; we call such a task the unintended task $U$.
Precisely, to solve the unintended task $U$, a consumer team trains their \textbf{\emph{consumer model}} $C$ on top of the produced embeddings.

The above setting is prevalent in industry, where the produced embeddings are widely shared within the organization to power a wide variety of unintended tasks~\citep{ying2018graph}.
As a concrete example, let us consider a recommender system application. The intended task $T$ could be user-item interaction prediction.
We can use a Graph Neural Network (GNN)~\citep{ying2018graph,hamilton2017inductive} as the embedding model $M$ to generate user and item embeddings, which are then used to produce the likelihood score that the user will interact with a given item.
At the same time these embeddings can be used by many consumer teams to perform their unintended tasks.
For instance, a consumer team can use the item embeddings to build model $C$ to solve the task $U$ of detecting fraudulent items.

\xhdr{Embedding model version updates}
The embedding team updates the embedding model $M$ every once in a while to improve the performance on the intended task $T$.
We use $M_0, M_1, M_2, \ldots$ to denote the evolving versions of the embedding model, where $M_k$ is the $k$-th model version.
At version $k$, we learn $M_k$ to solve $T$ by minimizing the objective:
\begin{align}
\label{eq:intended_obj}
    L_k(M_k).
\end{align}

Given single input data $x$, each \textbf{\emph{ver-$k$ embedding model}} $M_k$ produces \textbf{\emph{ver-$k$ embedding}} $\vb{z}_k \equiv M_k(x)$. 
The collection of the ver-$k$ embeddings is denoted as $\{\vb{z}_k\}$, which is computed over current sets of input data points and may be refreshed as new data arrives.
Moreover, for each version $k$, we consider a consumer team that uses ver-$k$ embeddings and consumer model $C_k$ to solve their unintended task $U_k$.

\xhdr{Compatibility of embeddings}
Different versions $j < k$ of the embeddings $\vb{z}_k, \vb{z}_j$ may be {\em incompatible} because of the difference between $M_k$ and $M_j$ that generate them.
This presents an issue that consumer model $C_j$ trained on ver-$j$ embeddings will not be compatible with the later ver-$k$ embeddings. Feeding $\vb{z}_k$ into $C_j$ will give random/arbitrary predictions.

To resolve this issue, in Section \ref{sec:method}, we develop a cost-efficient framework to generate \textbf{\emph{ver-$j$ compatible embedding}} $\widetilde{\vb{z}}_j$ from later-version embedding $\vb{z}_k$ such that feeding $\widetilde{\vb{z}}_j$ into $C_j$ gives robust predictive performance.
We consider the problem of \textbf{\emph{backward compatible embedding learning}}, \ie, learn to produce $\widetilde{\vb{z}}_j$ from $\vb{z}_k$ for any historical version $j < k$.

\subsection{Generality of our Setting}
\label{subsec:generality}
We show that the above simple problem setting is general enough to capture complex real-world scenarios.

\xhdr{Complex embedding model evolution} Our assumption on the evolving embedding model $M$ is minimal: Each model version $M_k$ just needs to output an embedding given an input data point. Our setting, therefore, allows different $M_k$'s to have different internal model architectures and output dimensionality.
For instance, our setting allows a later-version embedding model to use a more advanced model architecture.

\xhdr{Evolving training data and loss functions}
Not only can different $M_k$'s have different architectures, they can also be trained on different data and with different loss functions.
All such differences are absorbed into ver-$k$-specific objective $L_k$ in Eq.~\eqref{eq:intended_obj}.
For instance, our setting allows later-version embedding model to be trained on more data with an improved loss function.

\xhdr{Multiple different intended tasks}
We consider a single shared intended task $T$ for simplicity, but our setting naturally handles multiple intended tasks that are different for different embedding model versions.
This is because we only assume each $M_k$ to be trained on objective $L_k$, which can be designed to solve different intended tasks for different version $k$.

\xhdr{Multiple consumer teams}
For each version $k$, we consider a single consumer team solving $U_k$ using model $C_k$ for simplicity. However, our setting naturally handles \emph{multiple} consumer teams using the same embedding version $k$. We can simply index the consumer teams as solving $U_{k,0}, U_{k,1}, U_{k,2}, \ldots$  using models $C_{k,0}, C_{k,1}, C_{k,2}, \ldots$, respectively.

\section{Methodological Framework}
\label{sec:method}
Here we present our framework to learn backward compatible embeddings.
At version $k$, our framework mainly keeps the latest ver-$k$ embedding model $M_k$ and its generated embeddings $\{\vb{z}_k\}$; hence, we say our framework follows the \textbf{\emph{keep-latest}} (embedding model) approach.
We start by contrasting it with what we call \textbf{\emph{keep-all}} (embedding models).
\subsection{Ideal but Costly Baseline: Keep-All}
\label{subsec:keepall}
Given unlimited human and computational budget, we can simply \emph{keep all versions of the embedding model} $M_0, M_1, \ldots, M_{k}$ and let them produce embeddings $\{\vb{z}_0\}, \{\vb{z}_1\}, \ldots, \{\vb{z}_{k}\}$.
Then, for any $j \leq k$, $\{\vb{z}_j\}$ can be directly used by consumer model $C_j$ in a compatible manner. 
We refer to this approach as {\em keep-all} (embedding models), which we formalize below.

\xhdr{Training setting}
At version $k$, we learn each $M_k$ by minimizing objective $L_k$ of Eq.~\eqref{eq:intended_obj} to solve the intended task $T$.
Once $M_k$ is trained to produce ver-$k$ embeddings, a consumer team trains their model $C_k$ on them to solve their unintended task $U_k$.

\xhdr{Inference setting}
At version $k$, we keep all versions of the embedding model learned so far: $M_0, M_1, \ldots, M_k$.
To solve the intended task $T$, we use the latest-version embedding model $M_k$ to produce embeddings $\{\vb{z}_k\}$ and store them.
In addition, we use all the historical versions of the embedding model to produce the embeddings $\{\vb{z}_0\}, \{\vb{z}_1\}, \ldots, \{\vb{z}_{k-1}\}$ and store all of them.
For any $j \leq k$, we can perform unintended task $U_j$ by simply feeding compatible embeddings $\{\vb{z}_j\}$ to consumer model $C_j$.

\xhdr{Issues with Keep-All} 
The issue with the keep-all approach is that it is too costly in large-scale applications, \eg, web-scale recommender systems that utilize billions of embeddings~\citep{ying2018graph,zhu2019aligraph}. This is because embeddings need to be produced and stored for \emph{every} version.\footnote{We assume the inference cost of $M_k$ is high, which makes it costly to infer ver-$k$ embedding every time it is requested.} The cost of maintaining all versions of the embeddings and the embedding model quickly grows especially when the embedding model version is updated frequently.
Despite the impracticality, the keep-all approach sets the high standard in terms of the intended and unintended task performance, which we try to approximate with our cost-efficient framework.

\subsection{Our Framework: Keep-Latest}

Our framework follows the keep-latest approach, where only the latest-version embedding model and embeddings are kept at any given timestamp.
\vspace{0.02cm}

\xhdr{Backward transformation}
The key to our approach is to learn a \textbf{\emph{backward (compatibility) transformation}} $B_k: \mathbb{R}^{D_k} \to \mathbb{R}^{D_{k-1}}$ that aligns ver-$k$ embedding $\vb{z}_k \in \mathbb{R}^{D_k}$ to its previous version $\vb{z}_{k-1} \in \mathbb{R}^{D_{k-1}}$. 
$B_k$ has to be light-weight so that the alignment can be performed cheaply on-the-fly.
Whenever we update $M_{k-1}$ to $M_k$, we learn the backward transformation $B_k$ to align the output of $M_k$ back to that of $M_{k-1}$.
At version $k$, we maintain all the backward transformations learned so far: $B_1, \ldots, B_k$.
Maintaining $B_1, \ldots, B_k$ is much cheaper than maintaining and storing all historical versions of the embeddings $\{\vb{z}_0\}, \{\vb{z}_1\}, \ldots, \{\vb{z}_{k-1}\}$.

The key insight is that we can compose the backward transformations to align $\vb{z}_k$ into any of its historical version $j < k$.
Let us introduce the \emph{composed backward function} ${B}^{j}_k  \equiv {B}_{j+1} \circ \cdots \circ {B}_{k}$.
We see that $B_k^{k-1} \equiv B_k$ and ${B}^{j}_k$ aligns ver-$k$ embedding $\vb{z}_k$ to ver-$j$ embedding $\vb{z}_j$.
As the alignment may not be perfect in practice, we say ${B}^{j}_k$ transforms $\vb{z}_k$ into ver-$j$ \emph{compatible} embedding $\widetilde{\vb{z}}_j$:
\begin{align}
\label{eq:compose1}
    \widetilde{\vb{z}}_j = {B}^{j}_k \left(\vb{z}_k\right).
\end{align}
Our aim is to have $\widetilde{\vb{z}}_j \approx \vb{z}_j$ so that $\widetilde{\vb{z}}_j$ can be fed into $C_j$ to give robust predictive performance on unintended task $U_j$.

\xhdr{Function alignment}
We wish to use $B_k$ to align $\vb{z}_k\equiv M_k(x)$ to $\vb{z}_{k-1} \equiv M_{k-1}(x)$ for every $x$, which reduces to aligning two \emph{functions}: $B_k \circ M_k$ and $M_{k-1}$.
We introduce function alignment objective $L_{\rm align}(B_k \circ M_k, M_{k-1})$, which encourages the two functions to be similar, \ie, given same input, produce similar output. We discuss the specific realization of $L_{\rm align}$ in Section \ref{subsec:choice_loss_align}.

\label{subsec:keeplatest}
\xhdr{Training setting}
We propose to add the function alignment objective $L_{\rm align}$ to the original objective $L_k$ for solving $T$:
\begin{align}
\label{eq:keeplatest_intended_obj}
    L_k(M_k) + \lambda \cdot L_{\rm align} \left(B_k \circ M_k, M_{k-1} \right),
\end{align}
where $\lambda > 0$ is a trade-off hyper-parameter.
At version $k$, parameters of $M_k$ and ${B}_k$ are learned, and the parameters of the previous version $M_{k-1}$ are fixed.
Once $M_k$ and ${B}_k$ are learned, we can safely discard $M_{k-1}$ because $M_{k-1}$ can be approximately reproduced by $B_k \circ M_k$.

\xhdr{Inference setting}
At version $k$, we only need to maintain the latest-version embedding model $M_k$, and a series of transformation functions learned so far: $B_1, \ldots, B_k$.
Consider an ideal situation after minimizing Eq.~\eqref{eq:keeplatest_intended_obj}, where we have the perfect function alignment \emph{for every version} until version $k$.
Then, we have the following single-step equations:
\begin{align}
\label{eq:chain}
 B_k \circ M_k = M_{k-1}, \  B_{k-1} \circ M_{k-1} = M_{k-2}\ , \ldots,  \ B_1 \circ M_1 = M_0.
\end{align}

In this ideal situation, we see that the composed backward function in Eq.~\eqref{eq:compose1} can exactly reproduce \emph{any} historical version of the embedding model $M_j (j < k)$ as
\begin{align}
\label{eq:compose}
{B}^{j}_k \circ M_k = M_j.
\end{align}
In practice, each equation in Eq.~\eqref{eq:chain} only holds approximately, and the approximation error of Eq.~\eqref{eq:compose} increases for smaller $j$ or larger $k$ due to more accumulation of the single-step approximation errors in Eq.~\eqref{eq:chain}. In our experiments, however, we find that the approximation error stays relatively stable over time and only increases sub-linearly with larger $k$ (\ie, using later-version embedding model to approximate $M_0$).
In Section \ref{subsec:choice_loss_align}, we mathematically analyze the approximation error and provide a possible explanation for the robust approximation performance even after multiple rounds of model updates.

\xhdr{Inductive capability}
Eq.~\eqref{eq:compose}, or more realistically, ${B}^{j}_k \circ M_k \approx M_j$, implies that our framework is fully inductive.
Given unseen data $x$, we can first obtain its latest ver-$k$ embedding ${\vb z}_k$ and store it. Then, ver-$j$ compatible embedding $\widetilde{\vb{z}}_j$ for any $j$ can be quickly obtained by $B^{j}_k(\vb{z}_k)$ and is expected to be similar to the actual ver-$j$ embedding $\vb{z}_j \equiv M_j(x)$.
Importantly, $\widetilde{\vb{z}}_j$ is obtained on-the-fly without being stored nor requiring the past model $M_j$.
In our experiments, we utilize this inductive capability of $B_k^j$ to transform embeddings unseen during training.


\subsubsection{Choices of ${B}_{k}$}
\label{subsec:choice_Align}

We want backward transformation $B_k$ to be light-weight. Here we consider two natural choices.

\xhdr{Linear} We use ${B}_{k}({\vb z}_k) = {\vb W}_{k}{\vb z}_k$, where ${\vb W}_{k} \in \mathbb{R}^{D_{k-1} \times D_k}$ is a learnable weight matrix.
In this case, Eq.~\eqref{eq:compose1} is written as
\begin{align}
\label{eq:linear}
\widetilde{{\vb z}}_j = {\vb W}^j_{k} {\vb z}_k,
\end{align}
 where ${\vb W}^j_k \equiv {\vb W}_{j+1} \cdots  {\vb W}_{k} \in \mathbb{R}^{D_{j} \times D_{k}}$ is pre-computed for every $j$.

\xhdr{NoTrans} As a baseline, we also consider not applying any transformation to ${\vb z}_k$. In other words, our backward transformation functions are all identity functions. This means that when training $M_k$, the produced ver-$k$ embedding $\vb{z}_k$ is learned to be directly similar to its previous version $\vb{z}_{k-1}$. Therefore, $\vb{z}_k$ is directly backward compatible with $\vb{z}_{k-1}$ as well as any of its historical version $\vb{z}_{j}$. 
In the case of $D_k \geq D_{k-1}$, we simply take the first $D_{k-1}$ elements of ${\vb z}_k$.


\xhdr{Remark on limited expressiveness of NoTrans}
NoTrans seems like a very natural solution to our problem as it enforces $\vb{z}_k$ to be directly similar to $\vb{z}_{k-1}$ (\eg, by minimizing Euclidean distance between embeddings $\vb{z}_k$ and $\vb{z}_{k-1}$).
However, NoTrans is not desirable when the embeddings suitable for performing the intended task $T$ change over time as a result of distribution shift.
For instance, in recommender systems, users' interests and items' popularity change over time, so we want their embeddings to also change over time, which is discouraged in the NoTrans case.
In contrast, the Linear case allows $\vb{z}_k$ to be different from $\vb{z}_{k-1}$.
The additional expressiveness is crucial for $\vb{z}_k$ to perform well on the intended task $T$, as we will show in our experiments.


\subsubsection{Choices of $L_{\rm align}$ and Preventing Error Amplification}
\label{subsec:choice_loss_align}

\xhdr{Single-step alignment loss}
The role of the alignment objective $L_{\rm align}$ is to make $B_k \circ M_k$ similar to $M_{k-1}$.
We enforce the alignment on a set of data points $\mathcal{X} = \{x\}$, which we assume to be given. For instance, in recommender systems, $\mathcal{X}$ can simply be all the users and items.
Then, our alignment objective becomes:
\begin{align}
\label{eq:loss_align_single_step}
    L_{\rm align}\left(B_k \circ M_k, M_{k-1} \right)  
    & = \frac{1}{|\mathcal{X}|} \sum_{x \in \mathcal{X}} \left\| {B}_{k}\circ M_k(x) - M_{k-1}(x) \right\|^2 \nonumber \\
    & = \frac{1}{|\mathcal{X}|} \sum_{x \in \mathcal{X}} \left\| {B}_{k}\left( {\vb z}_k \right) - {\vb z}_{k-1} \right\|^2 \nonumber \\
    & = \frac{1}{|\mathcal{X}|} \sum_{x \in \mathcal{X}} \left\|
    {\bm \delta}_{k}(x) \right\|^2,
\end{align}
where ${\bm \delta}_{k}(x) \equiv {B}_{k}\left( {\vb z}_{k} \right) - {\vb z}_{k-1}$ is the \emph{single-step} alignment error between ${B}_{k}\left( {\vb z}_{k} \right)$ and $\vb{z}_{k-1}$ on a data point $x$. 

While natural, it is unclear how well the single-step alignment loss of Eq.~\eqref{eq:loss_align_single_step} enforces the small \emph{multi-step} alignment error ${\bm \delta}^{j}_{k}(x) \equiv {B}_{k}^j\left( {\vb z}_k \right) - {\vb z}_{j}$, where $j < k-1$.
Below we mathematically characterize their relation.

\xhdr{Error amplification}
Let us focus on the linear case of Eq.~\eqref{eq:linear}. Then, the multi-step alignment error on a single data point $x$ becomes:
\begin{align}
\label{eq:multi_step_error}
     {\bm \delta}^{j}_{k}(x)  =  {\vb W}^j_{k} {\vb z}_k - {\vb z}_j.
\end{align}
We note that Eq.~\eqref{eq:multi_step_error} \emph{cannot} be optimized directly because the keep-latest approach assumes $M_{j}$ and ${\vb z}_{j}$ are no longer kept when $M_{k}$ is being developed.
However, we learn from Eq.~\eqref{eq:loss_align_single_step} that all the historical backward transformation weights ${\vb W}_{1}, {\vb W}_{2}, \ldots, {\vb W}_{k-1}$, have been learned to minimize the L2 norm of the single-step alignment errors, ${\bm \delta}_{1}(x), {\bm \delta}_{2}(x), \ldots, {\bm \delta}_{k-1}(x)$, respectively.

We can rewrite the multi-step alignment error ${\bm \delta}^{j}_{k}(x)$ in Eq.~\eqref{eq:multi_step_error} using the single-step alignment errors, ${\bm \delta}_{j+1}(x), {\bm \delta}_{j+2}(x), \ldots, {\bm \delta}_{k}(x)$:
\begin{align}
\label{eq:calculation}
    {\bm \delta}^{j}_{k}(x) & = {\vb W}^{j}_{k-1} \left( {\vb W}_{k} {\vb z}_k \right) - {\vb z}_{j} \nonumber \\
    & = {\vb W}^{j}_{k-1} \left( {\vb z}_{k-1} + {\bm \delta}_{k}(x) \right) - {\vb z}_{j} \nonumber  \\
    & = \left\{ {\vb W}^{j}_{k-2} \left( {\vb W}_{k-1} {\vb z}_{k-1}  \right) - {\vb z}_{j} \right\} + {\vb W}^{j}_{k-1} {\bm \delta}_{k}(x)  \nonumber  \\
    & \qquad \qquad \qquad \qquad \vdots \nonumber \\
    & = \left(  {\vb W}^j_{j+1} {\vb z}_{j+1} - {\vb z}_{j} \right) + {\vb W}^j_{j+1} {\bm \delta}_{j+2}(x) + \cdots + {\vb W}^{j}_{k-1} {\bm \delta}_{k}(x) \nonumber \\
    & = {\bm \delta}_{j+1}(x) + {\vb W}^j_{j+1} {\bm \delta}_{j+2}(x) + \cdots + {\vb W}^{j}_{k-1} {\bm \delta}_{k}(x).
\end{align}
From Eq.~\eqref{eq:calculation}, we see that the single-step errors are not simply added up but are potentially \emph{amplified} by the historical backward transformation weights.
 We call this \textbf{\emph{error amplification}}. Minimizing the single-step error does not necessarily lead to the smaller amplified error, causing the large multi-step alignment error of Eq.~\eqref{eq:multi_step_error} in practice. This in turn deteriorates the unintended task performance.

\xhdr{Multi-step alignment loss}
To suppress the error amplification, here we develop the \emph{multi-step alignment loss}.
We see from the last term of Eq.~\eqref{eq:calculation} that the error ${\bm \delta}_{k}(x)$ made at version $k$ (learning $M_k$ and ${\vb W}_{k}$) gets amplified by ${\vb W}^{j}_{k-1}$ in the multi-step error ${\bm \delta}^{j}_{k}(x)$.
Our multi-step alignment loss explicitly suppresses the amplified error for every $j=0,\ldots, k-2, k-1$:
\begin{align}
\label{eq:loss_align_prevent_amplify}
   & \frac{1}{k} \cdot \bigg( \left\| {\vb W}^{0}_{k-1} {\bm \delta}_{k}(x) \right\|^2 + \cdots + \left\| {\vb W}^{k-2}_{k-1} {\bm \delta}_{k}(x) \right\|^2  + \left\|{\vb W}^{k-1}_{k-1} {\bm \delta}_{k}(x) \right\|^2  \bigg),
\end{align}
where we note that ${\vb W}^{k-1}_{k-1}=\vb{I}$. Thus, our multi-step alignment loss ensures the error ${\bm \delta}_{k}(x)$ made at version $k$ would not get amplified when we compute its historical version $\widetilde{\vb{z}}_j$ for any $j < k$.


The final multi-step alignment loss is the average of Eq.~\eqref{eq:loss_align_prevent_amplify} over $x\in \mathcal{X}$, analogous to Eq.~\eqref{eq:loss_align_single_step}.
Although Eq.~\eqref{eq:loss_align_prevent_amplify} contains $k$ terms, computing the loss itself is often much cheaper than computing embedding ${\vb z}_k$, so Eq.~\eqref{eq:loss_align_prevent_amplify} adds negligible computational cost compared to the single-step loss of Eq.~\eqref{eq:loss_align_single_step}.

\xhdr{Remark on the NoTrans case}
We note that NoTrans method does not suffer from the error amplification. To see this, we can replace all the weight matrices in Eq.~\eqref{eq:calculation} with identity matrices, resulting in the simple additive accumulation of the non-amplified errors.
However, NoTrans suffers from the limited expressiveness, as discussed in Section \ref{subsec:choice_Align}.

\xhdr{Remark on Error Accumulation}
As shown in Eq.~\eqref{eq:calculation}, both the Linear and NoTrans would suffer from the additive accumulation of the single-step errors.
There are $k-j$ single-step error terms in Eq.~\eqref{eq:calculation}. Assuming each error is an independent random variable with zero mean and the same standard deviation, we can conclude that the L2 norm of Eq.~\eqref{eq:calculation} only grows in the order of $\sqrt{k-j}$, \ie, sub-linearly w.r.t. the number of versions. In our experiments, we indeed observe that the L2 norm of the multi-step error grows gradually and sub-linearly  (Figure~\ref{fig:time_evolution}), and as a result, the unintended task performance stays robust even after multiple rounds of embedding version updates (Figure~\ref{fig:task_performance})

\subsubsection{Choices of Training Strategies of ${B}_{k}$}
\label{subsec:choice_opt}

We consider two strategies to train ${B}_{k}$ via Eq.~\eqref{eq:keeplatest_intended_obj}.

\xhdr{Joint-Align}
We jointly train ${B}_{k}$ with $M_k$ to minimize Eq.~\eqref{eq:keeplatest_intended_obj}.

\xhdr{Posthoc-Align}
We first train $M_k$ to minimize the $L_k$ objective. We then fix $M_k$ and train ${B}_{k}$ to minimize the $L_{\rm align}$ objective in a post-hoc manner.

\subsubsection{Method Variants}
\label{subsec:design_choice}

In all, we explore six different method variants under our framework, as shown in Table~\ref{tab:method_taxonomy}, each adopting different design choices introduced in the previous subsections.
We note that the techniques used in some method variants were already presented by prior works in different contexts.
Specifically, the Joint-NoTrans was originally presented by \citep{shen2020towards} in the context of backward compatible representation learning in open-set image recognition. The Posthoc-Lin-SLoss is broadly adopted in cross-lingual word embedding alignment~\citep{mikolov2013exploiting}. We will further discuss this in Section \ref{sec:related}.

Empirically, we will show that these two variants are outperformed by the best method under our framework, namely Joint-Lin-MLoss. We give a special name to this method, \method.

\begin{table}[t]
    \centering
        \caption{A set of all 6 method variants we consider under our framework. We vary transformation function (linear vs. no transformation), alignment loss function (single- vs. multi-step) and alignment (joint vs. posthoc)\protect\footnotemark .}
    \resizebox{\columnwidth}{!}{%
    {\fontsize{8}{8}\selectfont
    \begin{tabular}{c|cc}
      \toprule
       \textbf{Trans function} / \textbf{$\vb{L}_{\rm align}$} & \textbf{Joint-Align} & \textbf{Posthoc-Align}\\
       \midrule
         Linear / Single-Step-Loss & Joint-Lin-SLoss & Post-Lin-SLoss~\citep{mikolov2013exploiting} \\
         \mr{2}{Linear / Multi-Step-Loss} & Joint-Lin-MLoss & \mr{2}{Post-Lin-MLoss} \\
          & \textbf{(BC-Aligner)} & \\
      \midrule
        NoTrans / Single-Step-Loss & Joint-NoTrans~\citep{shen2020towards} & Non-BC \\
      \bottomrule
    \end{tabular}}}
    \label{tab:method_taxonomy}
  \vspace{-3mm}
\end{table}

\footnotetext[3]{The NoTrans model does not require the multi-step alignment loss. The combination of the NoTrans model and the Posthoc-Align does \emph{not} guarantee backward compatibility, as the identity has no parameter to learn in the post-hoc alignment stage; hence, we call it ``Non-BC''.}    

\section{Evaluation Framework}
\label{sec:eval}
Here we present an evaluation framework to measure the success of the keep-latest approach presented in Section \ref{sec:method}.
We consider a series of embedding model updates, $M_0, \cdots, M_K$ to improve the performance of the intended task $T$.
At the same time, consumer teams train their models to solve their unintended tasks.
Without loss of generally, we only consider a consumer team that uses ver-0 embeddings (generated by $M_0$) to solve unintended task $U_0$.
In our experiments, we consider multiple such consumer teams as discussed in Section \ref{subsec:generality}.

We provide three summary metrics calculated at every version $k = 0, 1, \ldots, K$. 
To make the comparisons meaningful, we assume the keep-all and keep-latest approaches share the same base objective $L_k$ and model architecture $M_k$ for every $k$.

\xhdr{(1)  Degradation of intended task performance compared to keep-all}
The keep-all approach provides an upper bound in terms of the intended task $T$ performance, as $M_k$ is solely optimized for $L_k$. Note, however, that this upper bound is impractical to achieve in most settings, as the keep-all approach is prohibitively costly. 
The keep-latest approach (and potentially other approaches) needs to maintain backward compatibility in addition to optimizing for $L_k$, which could deteriorate its intended task performance.
Therefore we measure the intended task performance \emph{degradation} compared to $M_k$ trained solely with $L_k$. 

\xhdr{(2) Degradation of unintended task performance compared to keep-all}
The keep-all approach also provides an upper bound in terms of the unintended task $U_0$ performance. This is because consumer model $C_0$ is optimized to perform well on ver-0 embeddings, which can be directly produced by the keep-all approach via kept $M_0$.
On the other hand, at version $k$, the keep-latest approach does not have access to $M_0$ and can only approximate $\vb{z}_0$ by backward compatible embedding $\widetilde{\vb{z}}_0$.
Therefore, we measure the unintended task $U_0$ performance \emph{degradation} when using ver-0 compatible embedding $\widetilde{\vb{z}}_0$ as opposed to using the actual ver-0 embedding $\vb{z}_0$.

\xhdr{(3) Embedding alignment error}
Metric (2) above is dependent on the choice of the unintended task $U_0$, which may not cover the entire spectrum of unintended tasks for which the embeddings can be used. 
It is therefore useful to have task-agnostic metric that generally correlates well with a wide range of unintended tasks.
To this end, we propose to measure the embedding alignment error between $\widetilde{\vb{z}}_0$ (\eg, we use ${B}_1\circ \cdots \circ {B}_k \left(\vb{z}_k\right)$ in our methods) and the actual $\vb{z}_0$.
We calculate it as the L2 distance between $\widetilde{\vb{z}}_0$ and $\vb{z}_0$, which we then average over a set of data points.

\section{Experiments}
\label{sec:experiments}

We evaluate our methods following the evaluation protocol presented in Section \ref{sec:eval}. 
We start with introducing a new benchmark in Section \ref{subsec:recsys_benchmark}. Then, we present experimental results in Section \ref{subsec:results}.

\begin{table}
    \centering
    \caption{{\bf Statistics of 3 different dynamic datasets we use.}
    }
    \label{tab:dataset_stats}
    \resizebox{\columnwidth}{!}{%
    {\fontsize{8}{8}\selectfont
    \begin{tabular}{lrrrrr}
      \toprule
        \mr{2}{\textbf{Dataset}} & \mc{3}{c}{} & \mc{2}{c}{\textbf{Feature stats}} \\
         & \#Users & \#Items & \#Interact. & \#Brands & \#Sub cat. \\
      \midrule
        Musical Instruments      & 27,530 & 10,611 & 231,312 & 391 & 349  \\
        Video Games  & 55,223 & 17,389  & 496,315  & 330 & 149 \\
        Grocery       & 127,496 & 41,280 & 1,143,063 & 1,806 & 774 \\
      \bottomrule
    \end{tabular}}}
    \vspace{-3mm}
\end{table}

\subsection{Recommender System Benchmark}
\label{subsec:recsys_benchmark}
\begin{table*}
    \centering
    \caption{Results over 3 dataset. Absolute performance of the Keep-All is included in brackets. For all the metrics, the relative degradation is computed after the average is taken across different timestamps and the five unintended tasks. We see from the 3rd column that the \method{} provides the best trade-off between the intended and unintended task performance, yielding the closest performance to the costly Keep-All approach. }
    \vspace{-2mm}
    \label{tab:summary-table}
    \renewcommand{\arraystretch}{1.1}
    {\fontsize{8}{8}\selectfont
    \begin{tabular}{lllcccc}
      \toprule
        \mr{3}{\textbf{Dataset}} & \mr{3}{\textbf{Approach}} & \mr{3}{\textbf{Method}} & \textbf{(1) Intented task} & \textbf{(2) Unintended task} & \textbf{(1)+(2)} & \textbf{(3) Emb}\\
d         & & & \textbf{Degradation} from & \textbf{Degradation} from & \textbf{Degradation} from & \textbf{Align}  \\
         & & & Keep-All (\%) $\uparrow$ & Keep-All (\%) $\uparrow$ & Keep-All (\%) $\uparrow$ & \textbf{Error} $\downarrow$ \\
      \midrule
        & Keep-All & Keep-All (Abs. perf.) & -0.00 (12.13) & -0.00 (68.66) & -0.00 & 0.00 \\  \cline{2-7}
        & \mr{2}{Keep-$M_0$} & Fix-$M_0$ & -26.81 & -0.00 & -26.81 & 0.00 \\
        &  & Finetune-$M_0$~\citep{goyal2018dyngem} & -7.69 & -7.46 & -15.15 & 1.01 \\  \cline{2-7}
        & \mr{6}{Keep-Latest} & Non-BC & -0.00 & -26.45 & -26.45 & 2.61 \\
        Musical & & Post-Lin-SLoss~\citep{mikolov2013exploiting} & -0.00 & -10.25 & -10.25 & 1.27 \\
        Instruments & & Post-Lin-MLoss & -0.00 & -14.43 & -14.43 & 1.34 \\ 
        & & Joint-NoTrans~\citep{shen2020towards} & -9.00 & -0.80 & -9.80 & 0.41 \\
        & & Joint-Lin-SLoss & -3.67 & -1.07 & -4.74 & 0.48 \\
        & & \textbf{\method{}} & -2.96 & -0.65 & \textbf{-3.62} & \textbf{0.38} \\
    \midrule
        & Keep-All & Keep-All (Abs. perf.) & -0.00 (12.69) & -0.00 (72.76) & -0.00 & -0.00 \\ \cline{2-7}
        & \mr{2}{Keep-$M_0$} & Fix-$M_0$ & -25.55 & -0.00 & -25.55 & 0.00 \\
        &  & Finetune-$M_0$~\citep{goyal2018dyngem} & -10.61 & -6.72 & -17.32 & 0.87 \\  \cline{2-7}
        & \mr{6}{Keep-Latest} & Non-BC & -0.00 & -30.81 & -30.81 & 2.65 \\
        Video & & Post-Lin-SLoss~\citep{mikolov2013exploiting} & -0.00 & -8.25 & -8.25 & 1.10 \\
        Games & & Post-Lin-MLoss & -0.00 & -14.56 & -14.56 & 2.51 \\ 
        & & Joint-NoTrans~\citep{shen2020towards} & -9.12 & -0.62 & -9.74 & 0.32 \\
        & & Joint-Lin-SLoss & -3.98 & -1.03 & -5.01 & 0.40 \\
        & & \textbf{\method{}} & -3.35 & -0.59 & \textbf{-3.94} & \textbf{0.28} \\
    \midrule
      \mr{9}{Grocery} & Keep-All & Keep-All (Abs. perf.) & -0.00 (7.78) & -0.00 (65.82) & -0.00 & -0.00 \\ \cline{2-7}
         & \mr{2}{Keep-$M_0$} & Fix-$M_0$ & -27.79 & -0.00 & -27.79 & 0.00 \\
        & & Finetune-$M_0$~\citep{goyal2018dyngem} & -24.74 & -5.52 & -30.26 & 1.12 \\  \cline{2-7}
        & \mr{6}{Keep-Latest} & Non-BC & -0.00 & -22.07 & -22.07 & 2.69 \\
        & & Post-Lin-SLoss~\citep{mikolov2013exploiting} & -0.00 & -6.36 & -6.36 & 1.30 \\
        & & Post-Lin-MLoss & -0.00 & -15.69 & -15.69 & 4.84 \\ 
        & & Joint-NoTrans~\citep{shen2020towards} & -11.31 & -0.33 & -11.64 & 0.18 \\
        & & Joint-Lin-SLoss & -2.90 & -2.17 & -5.07 & 0.52 \\
        & &\textbf{\method{}} & -3.21 & -0.07 & \textbf{-3.28} & \textbf{0.13} \\
      \bottomrule
    \end{tabular}}
    \vspace{-2mm}
\end{table*}

\subsubsection{Overview}
We consider evolving user-item bipartite graph datasets in recommender systems, where new users/items and their interactions appear over time.
As the intended task, we consider the standard user-item link prediction task, and use GNNs trained on this task as the embedding model that generate user/item embeddings. To simulate the phenomenon of increasing dataset sizes and model capabilities over time, we consider larger GNN models trained on more edges over time.
We consider five different unintended tasks that are of interest to practitioners. 
Below we explain the benchmark in more details.

\begin{figure*}[t]
  \centering
  \includegraphics[width=0.95\linewidth]{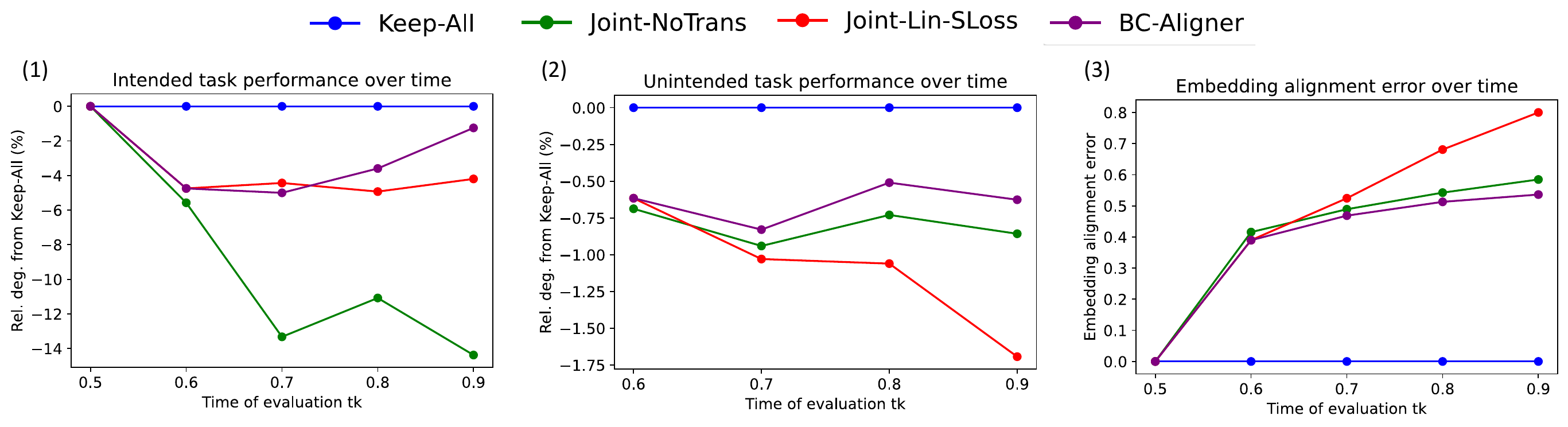}
  \vspace{-2mm}
  \caption{Performance over time. For the sub-figures (1) and (2), we plot the relative performance degradation from Keep-All in the $y$-axis (closer to zero, the better).}
  \label{fig:time_evolution}
  \vspace{-2mm}
\end{figure*}

\begin{figure}
  \centering
  \includegraphics[width=0.9\linewidth]{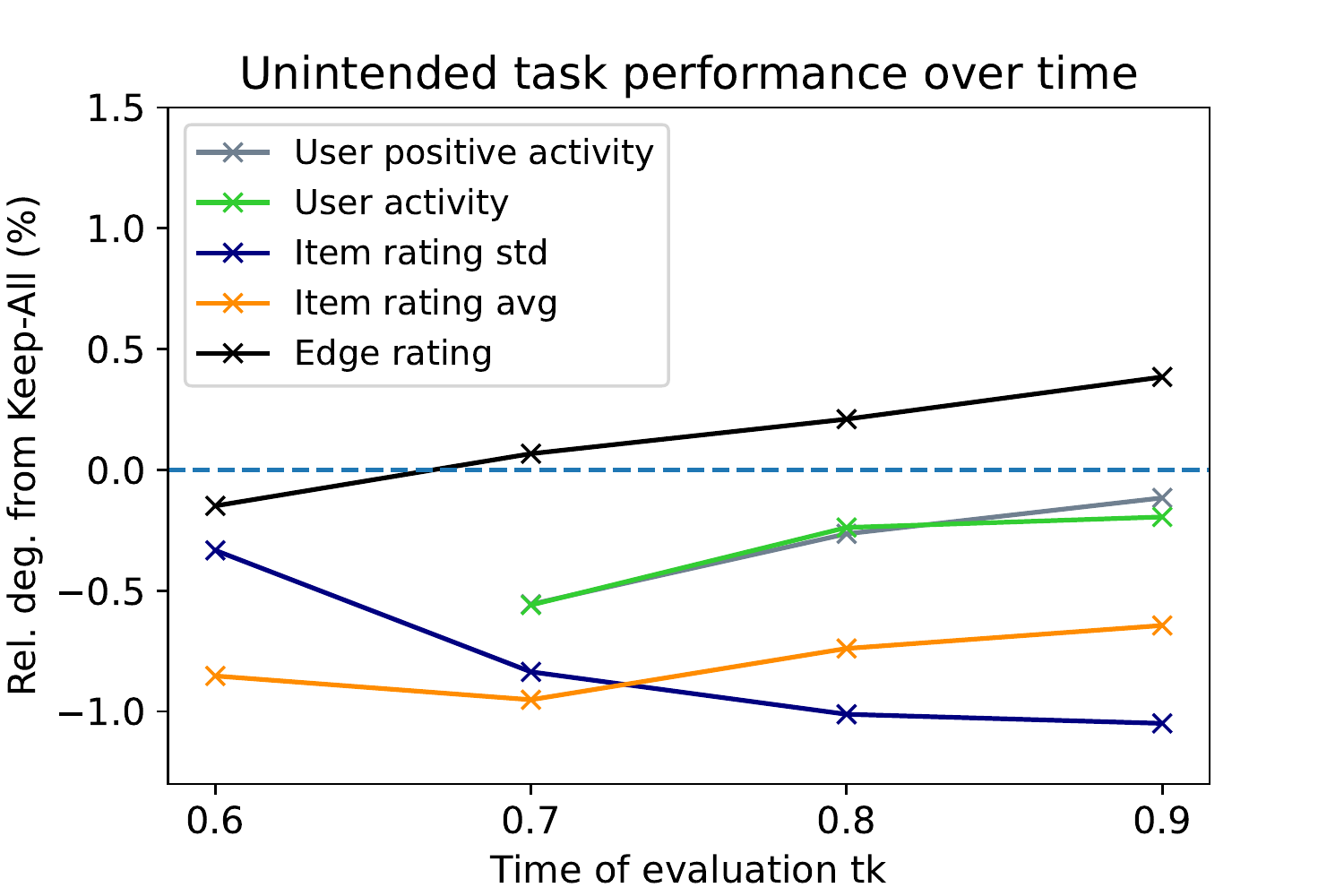}
  \caption{Unintended task performance degradation of \method{} compared to Keep-All (dotted line) over time.  The performance degradation of each unintended task stays relatively stable over time.
  }
  \label{fig:task_performance}
  \vspace{-2mm}
\end{figure}

\subsubsection{Datasets}
\label{subsubsec:datasets}
We use public Amazon Product Reviews dataset\footnote{Available at \url{https://jmcauley.ucsd.edu/data/amazon/}} that contains timestamped Amazon product reviews, spanning May 1996 - July 2014~\citep{ni2019justifying}. The entire dataset is partitioned according to the categories of products/items. In our experiments, we use the datasets from three categories: Musical Instruments, Video Games, and Grocery.
Each item has brand and subcategory features, that can be expressed as multi-hot vectors. User features are not available in this dataset. 
Dataset statistics is summarized in Table \ref{tab:dataset_stats}.

The datasets can be naturally modeled as evolving graphs, where nodes represent users/items, and each edge with a timestamp represents a user-item interaction, \ie, a user review an item at the particular timestamp. 
We scale the timestamp between 0 and 1, representing the ratio of edges observed so far, \eg, a timestamp of 0.6 means 60\% of edges have been observed until that timestamp. This allows us to compare methods across the datasets. 

We consider embedding models to be updated at $t_0 = 0.5, t_1 = 0.6, ... t_{K} = 0.9$, $K = 4$. In total, five versions of the embedding model are developed. We use $E_k$ to denote the set of all edges up to $t_k$.


\subsubsection{Intended Task}
\label{subsubsec:intended}
As the intended task, we consider the standard link prediction in recommender systems. 
\citep{wang2019neural}.
At every timestamp $t_k$ for $k=0,\ldots,K$, we train $M_k$ on $E_k$ and use it to predict on the edges between time $t_k$ and $t_{k+1}$, \ie, $E_{k+1} \setminus E_k$, where $t_{K+1}=1$ by construction.
We follow the same strategy as \citep{he2017neural,wang2019neural} to train and evaluate $M_k$.
Specifically, given user/item embeddings generated by $M_k$, we use the dot product to score the user-item interaction and use the BPR loss \citep{rendle2012bpr} to train $M_k$. We then evaluate $M_k$ on $E_{k+1} \setminus E_k$ using Recall@50.

\subsubsection{Embedding Models}
\label{subsubsec:emb_models}
We use the GraphSAGE models~\citep{hamilton2017inductive} as our core embedding models.
To simulate the updates in model architecture over time, we start with a small GraphSAGE mode at $t_0$ and make it larger (both deeper and wider) over time.
Please refer to Appendix \ref{app:emb_model} for more details.

\subsubsection{Methods and Baselines}
We consider the six methods under our framework, as depicted in Table~\ref{tab:method_taxonomy}. They all follow the cost-efficient Keep-Latest approach. For the joint-training methods, we set $\lambda=16$ in Eq.~\eqref{eq:keeplatest_intended_obj} unless otherwise specified.

We also consider the following two cost-efficient baseline methods that only keep the ver-0 embedding model $M_0$ (as opposed to the latest embedding model); we group the methods under \textbf{Keep-$M_0$}.

\xhdr{Fix-$M_0$}
We train $M_0$ at timestamp $t_0$ and fix its parameters throughout the subsequent timestamps.
For all the timestamps, the same $M_0$ is used to perform the link prediction task as well as to generate the embeddings for the unintended tasks.  
As ver-0 embeddings are always produced, there is no backward compatibility issue for Fix-$M_0$.
However, the method always uses $M_0$ and cannot benefit from additional data and better model architectures available in the future. This will impact performance on the intended task $T$. 

\xhdr{Finetune-$M_0$}
We also consider a method more advanced than the Fix-$M_0$, originally introduced by \citep{goyal2018dyngem}.
Specifically, we train $M_0$ at timestamp $t_0$. Then, in the subsequent timestamp $t_k$ with $1 \leq k$, we finetune $M_{k-1}$ to obtain $M_k$. Note that fine-tuning allows $M_0$ to learn from more data over time. However, the approach cannot benefit from improved model architecture over time, as fine-tuning is not possible for different model architectures.

We evaluate the methods against the costly Keep-All approach by measuring the performance degradation and the embedding alignment error explained in Section \ref{sec:eval}.
We specifically consider the relative performance degradation in \%, which is calculated as $\frac{100 \cdot \left( {\rm Perf} - {\rm Perf}_{\rm Keep-All} \right)}{{\rm Perf}_{\rm Keep-All}}$, where ${\rm Perf}$ is the performance of a method of interest. The value should be negative in most cases (as ${\rm Perf}_{\rm Keep-All}$ is the upper bound); the closer to zero, the better.
The trained $M_0$ is exactly the same across all the methods, as all the methods train $M_0$ using only $L_0$ and the same random seed.

Note that at every timestamp, we encounter new data (\eg, existing users/items with more interactions, new users/items), which are never seen at previous timestamps. As our embedding model and transformations are inductive, we can generate backward compatible embeddings $\widetilde{\vb{z}}_0$ on the new data and make its prediction.

\subsubsection{Unintended Tasks}
\label{subsubsec:unintended}
We prepare five unintended binary classification tasks that include prediction on users, items, and user-item interactions. 
We utilize the review rating information in designing the unintended tasks, which is not used by the intended link prediction task. 
For all the tasks, we use 1-hidden-layer MLPs and evaluate the performance using ROC-AUC.
Refer to Appendix~\ref{app:detail_unintended} for details.
\begin{itemize}
    \item \textbf{User activity prediction:} Predict whether a given user will interact with at least a single item in the near future.
    
    \item \textbf{User positive activity prediction:} Predict if a given user will have at least one positive interaction in the near future.
    
    \item \textbf{Item rating avg prediction:} Predict whether the average rating of a given item until the near future will be above a threshold.
    
    \item \textbf{Item rating std prediction:} Predict whether the standard deviation of the ratings of a given item until the near future will be above a threshold.
    
    \item \textbf{Edge rating prediction:} Predict whether a given user will give a positive rating to a given item.
\end{itemize}

\begin{figure*}
  \centering
  \includegraphics[width=0.95\linewidth]{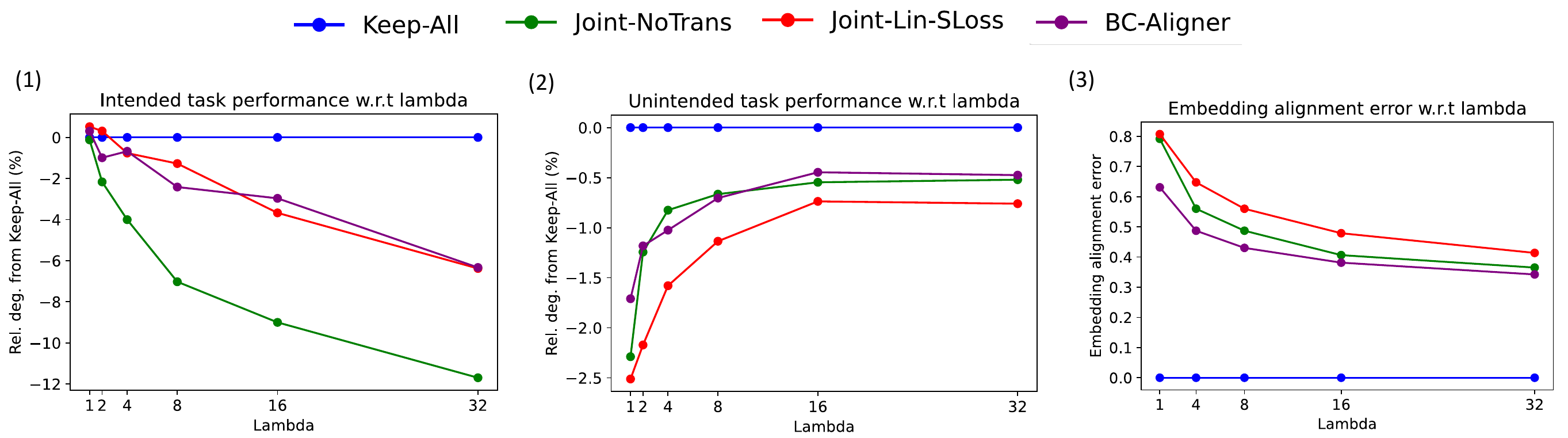}
  \vspace{-2mm}
  \caption{Performance as a function of the trade-off hyper-parameter $\lambda$. For sub-figures (1) and (2), we plot the relative degradation from Keep-All (closer to zero, the better).}
  \label{fig:lambda_sensitivity}
  \vspace{-2mm}
\end{figure*}

\subsection{Results}
\label{subsec:results}
In Table \ref{tab:summary-table}, we summarize the averaged results of different methods on the three metrics presented in Section \ref{sec:eval}, namely, (1) intended task degradation, (2) unintended task degradation, and (3) alignment error. 
We use the simple addition of intended task degradation and unintended task degradation as the unified metric to capture the trade-off between the intended and unintended task performance.

First, we observe high correlation between the unintended task degradation and alignment error; the smaller the unintended task performance degradation is, the smaller the embedding alignment error is.
This validates our claim that alignment error can be used as a general proxy for unintended task performance degradation.

Second, Non-BC suffers from large unintended task degradation and alignment error, as the new versions of the embedding model are not trained to be backward compatible.

Third, we see that Fix-$M_0$ suffers from large intended task degradation, indicating that it is highly sub-optimal to not update the embedding model over time to improve the intended task performance.
The more advanced Finetune-$M_0$ \citep{goyal2018dyngem} still suffers from large intended task degradation due to its inability of adopting the new model architectures over time. Moreover, the unintended task performance of Finetune-$M_0$ degrades by 5--7\%, implying that finetuned embedding models are generally \emph{no longer} compatible with the original model.

Fourth, Post-Lin-SLoss~\citep{mikolov2013exploiting} performs poorly on unintended task, degrading the performance by 6--10\%. This is likely due to the large embedding alignment error.

Fifth, Joint-NoTrans~\citep{shen2020towards} provides the small unintended task degradation and alignment error, but falls short on the intended task, degrading its performance by 9--11\%.
Compared to Joint-NoTrans, Joint-Lin-SLoss produces gives smaller intended task degradation. 
However, Joint-Lin-SLoss performs relatively poorly on unintended tasks, degrading the performance by 1--2\%.
As we will see in Figure \ref{fig:time_evolution}, this is possibly due to the amplification of the single-step embedding alignment error.

Overall, \method{} provides the best trade-off between the intended and unintended task performance, only suffering from around 3\% (resp.~0.5\%) degradation in the intended (resp.~unintended) task performance. It also achieves the smallest embedding approximation error among all the methods.

From now on, we consider the Musical Instruments dataset for all results. We focus on the three methods that give the most promising averaged results: Joint-NoTrans, Joint-Lin-SLoss, and \method{}.

\xhdr{Results over time} Figure \ref{fig:time_evolution} shows the intended task degradation, unintended task degradation, and alignment error over time.
For all the metrics, we observe that \method{} provides the best performance \emph{across almost all the timestamps}.
In Figure \ref{fig:time_evolution} (3), we see the sharp increase in the embedding alignment error over time for Joint-Lin-Sloss. This is likely due to the error amplification, as the single-step error (the error at $t=0.6$) is comparable across the methods. Indeed, once the multi-step alignment loss is used in \method{}, the embedding alignment error increases less sharply.

\xhdr{Results on each unintended task performance over time}
Figure \ref{fig:task_performance} shows the performance degradation of each unintended task over time, when \method{} is used. We see that the degradation from Keep-All is relatively stable over time.

\xhdr{Averaged results with varying $\lambda$}
Figure \ref{fig:lambda_sensitivity} shows how the trade-off parameter $\lambda$ in Eq.~\eqref{eq:keeplatest_intended_obj} affects the three metrics.
We consider $\lambda \in \{1, 2, 4, 8, 16, 32\}$, and the results are averaged over timestamps and unintended tasks.
As we increase $\lambda$, all the methods have larger degradation in the intended task performance, smaller degradation in the unintended task performance, and smaller embedding alignment error, as expected.
For fixed $\lambda$, we see that \method{} often gives the best or comparable performance compared to the other methods.
In practice, $\lambda$ should be chosen based on the trade-off one wants to achieve between the intended and unintended task performance.

\section{Related Work}
\label{sec:related}
\xhdr{Backward compatible representation learning}
Our problem formulation shares a similar motivation as  \citep{shen2020towards}, which considers backward compatible representation learning for open-set image recognition. 
\citet{shen2020towards,meng2021learning} update the embedding model so that embeddings computed by the updated model are directly comparable to those generated by the previous model. Our work differs from this work in two aspects. First, \citet{shen2020towards} only considers a single task of interest (face recognition), while our work considers a more practical scenario of having both intended and unintended tasks and evaluates the trade-offs between the two.
Second, \citet{shen2020towards} mainly consider single-step backward compatibility, while we focus on multi-step backward compatibility. We show that our novel multi-step alignment loss achieves very small degradation of unintended task performance even after multiple version updates.

\xhdr{Embedding alignment}
Our work builds on cross-lingual embedding alignment methods, where source word embeddings are aligned to target embeddings by learning a linear transformation function~\citep{mikolov2013distributed,klementiev2012inducing,ruder2019survey,glavas2019properly,xing2015normalized,artetxe2016learning, yao2018dynamic}. 
\citet{tagowski2021embedding} applies the embedding alignment technique to the graph domain, where they align a set of node2vec embeddings~\citep{grover2016node2vec} learned over different snapshots of an evolving graph. However, all these methods assume the embeddings are fixed, which could result in a large alignment error if two sets of pretrained embeddings are very distinct~\citep{zhang2019girls}. Unlike these methods, we jointly learn the embeddings along with the backward transformation function, achieving much better alignment performance and better unintended task performance.


\section{Conclusion}
\label{sec:conclusion}

In this paper, we formulated the practical problem of learning backward compatible embeddings.
We presented a cost-efficient framework to achieve the embedding backward compatibility even after multiple rounds of updates of the embedding model version.
Under the framework, we proposed a promising method, \method{}, that achieves a better trade-off between the intended and unintended task performance compared to prior approaches.
There are numerous future directions to investigate. 
For instance, the trade-off could be further improved by using more expressive backward transformation functions with non-linearity.
It is also of interest to assume some partial knowledge about the unintended tasks (\eg, the pre-trained consumer models are accessible to the embedding team) to actually \emph{improve} the unintended task performance without re-training the consumer models.
Finally, it is useful to apply our framework to other applications domains, such as those involving sentence and image embeddings.

\section*{Acknowledgements}
We thank Andrew Wang and Rex Ying for their early discussion on the work.
Weihua Hu is supported by Funai Overseas Scholarship and Masason Foundation Fellowship. 
We also gratefully acknowledge the support of
DARPA under Nos. HR00112190039 (TAMI), N660011924033 (MCS);
ARO under Nos. W911NF-16-1-0342 (MURI), W911NF-16-1-0171 (DURIP);
NSF under Nos. OAC-1835598 (CINES), OAC-1934578 (HDR), CCF-1918940 (Expeditions), 
NIH under No. 3U54HG010426-04S1 (HuBMAP),
Stanford Data Science Initiative, 
Wu Tsai Neurosciences Institute,
Amazon, Docomo, Hitachi, Intel, JPMorgan Chase, Juniper Networks, KDDI, NEC, Toshiba, and UnitedHealth Group.

The content is solely the responsibility of the authors and does not necessarily represent the official views of the funding entities.

\bibliographystyle{ACM-Reference-Format}
\bibliography{acmart}


\begin{thebibliography}{24}


\ifx \showCODEN    \undefined \def \showCODEN     #1{\unskip}     \fi
\ifx \showDOI      \undefined \def \showDOI       #1{#1}\fi
\ifx \showISBNx    \undefined \def \showISBNx     #1{\unskip}     \fi
\ifx \showISBNxiii \undefined \def \showISBNxiii  #1{\unskip}     \fi
\ifx \showISSN     \undefined \def \showISSN      #1{\unskip}     \fi
\ifx \showLCCN     \undefined \def \showLCCN      #1{\unskip}     \fi
\ifx \shownote     \undefined \def \shownote      #1{#1}          \fi
\ifx \showarticletitle \undefined \def \showarticletitle #1{#1}   \fi
\ifx \showURL      \undefined \def \showURL       {\relax}        \fi
\providecommand\bibfield[2]{#2}
\providecommand\bibinfo[2]{#2}
\providecommand\natexlab[1]{#1}
\providecommand\showeprint[2][]{arXiv:#2}

\bibitem[\protect\citeauthoryear{Artetxe, Labaka, and Agirre}{Artetxe
  et~al\mbox{.}}{2016}]%
        {artetxe2016learning}
\bibfield{author}{\bibinfo{person}{Mikel Artetxe}, \bibinfo{person}{Gorka
  Labaka}, {and} \bibinfo{person}{Eneko Agirre}.}
  \bibinfo{year}{2016}\natexlab{}.
\newblock \showarticletitle{Learning principled bilingual mappings of word
  embeddings while preserving monolingual invariance}. In
  \bibinfo{booktitle}{\emph{Conference on Empirical Methods in Natural Language
  Processing (EMNLP)}}. \bibinfo{pages}{2289--2294}.
\newblock


\bibitem[\protect\citeauthoryear{Glavas, Litschko, Ruder, and Vulic}{Glavas
  et~al\mbox{.}}{2019}]%
        {glavas2019properly}
\bibfield{author}{\bibinfo{person}{Goran Glavas}, \bibinfo{person}{Robert
  Litschko}, \bibinfo{person}{Sebastian Ruder}, {and} \bibinfo{person}{Ivan
  Vulic}.} \bibinfo{year}{2019}\natexlab{}.
\newblock \showarticletitle{How to (properly) evaluate cross-lingual word
  embeddings: On strong baselines, comparative analyses, and some
  misconceptions}.
\newblock \bibinfo{journal}{\emph{arXiv preprint arXiv:1902.00508}}
  (\bibinfo{year}{2019}).
\newblock


\bibitem[\protect\citeauthoryear{Goyal, Kamra, He, and Liu}{Goyal
  et~al\mbox{.}}{2018}]%
        {goyal2018dyngem}
\bibfield{author}{\bibinfo{person}{Palash Goyal}, \bibinfo{person}{Nitin
  Kamra}, \bibinfo{person}{Xinran He}, {and} \bibinfo{person}{Yan Liu}.}
  \bibinfo{year}{2018}\natexlab{}.
\newblock \showarticletitle{Dyngem: Deep embedding method for dynamic graphs}.
\newblock \bibinfo{journal}{\emph{arXiv preprint arXiv:1805.11273}}
  (\bibinfo{year}{2018}).
\newblock


\bibitem[\protect\citeauthoryear{Grover and Leskovec}{Grover and
  Leskovec}{2016}]%
        {grover2016node2vec}
\bibfield{author}{\bibinfo{person}{Aditya Grover} {and} \bibinfo{person}{Jure
  Leskovec}.} \bibinfo{year}{2016}\natexlab{}.
\newblock \showarticletitle{node2vec: Scalable feature learning for networks}.
  In \bibinfo{booktitle}{\emph{ACM SIGKDD Conference on Knowledge Discovery and
  Data Mining (KDD)}}. ACM, \bibinfo{pages}{855--864}.
\newblock


\bibitem[\protect\citeauthoryear{Hamilton, Ying, and Leskovec}{Hamilton
  et~al\mbox{.}}{2017}]%
        {hamilton2017inductive}
\bibfield{author}{\bibinfo{person}{William~L Hamilton}, \bibinfo{person}{Rex
  Ying}, {and} \bibinfo{person}{Jure Leskovec}.}
  \bibinfo{year}{2017}\natexlab{}.
\newblock \showarticletitle{Inductive Representation Learning on Large Graphs}.
  In \bibinfo{booktitle}{\emph{Advances in Neural Information Processing
  Systems (NeurIPS)}}. \bibinfo{pages}{1025--1035}.
\newblock


\bibitem[\protect\citeauthoryear{He, Liao, Zhang, Nie, Hu, and Chua}{He
  et~al\mbox{.}}{2017}]%
        {he2017neural}
\bibfield{author}{\bibinfo{person}{Xiangnan He}, \bibinfo{person}{Lizi Liao},
  \bibinfo{person}{Hanwang Zhang}, \bibinfo{person}{Liqiang Nie},
  \bibinfo{person}{Xia Hu}, {and} \bibinfo{person}{Tat-Seng Chua}.}
  \bibinfo{year}{2017}\natexlab{}.
\newblock \showarticletitle{Neural collaborative filtering}. In
  \bibinfo{booktitle}{\emph{www}}. \bibinfo{pages}{173--182}.
\newblock


\bibitem[\protect\citeauthoryear{Hu, Fey, Zitnik, Dong, Ren, Liu, Catasta, and
  Leskovec}{Hu et~al\mbox{.}}{2020}]%
        {hu2020open}
\bibfield{author}{\bibinfo{person}{Weihua Hu}, \bibinfo{person}{Matthias Fey},
  \bibinfo{person}{Marinka Zitnik}, \bibinfo{person}{Yuxiao Dong},
  \bibinfo{person}{Hongyu Ren}, \bibinfo{person}{Bowen Liu},
  \bibinfo{person}{Michele Catasta}, {and} \bibinfo{person}{Jure Leskovec}.}
  \bibinfo{year}{2020}\natexlab{}.
\newblock \showarticletitle{Open graph benchmark: Datasets for machine learning
  on graphs}. In \bibinfo{booktitle}{\emph{Advances in Neural Information
  Processing Systems (NeurIPS)}}.
\newblock


\bibitem[\protect\citeauthoryear{Kingma and Ba}{Kingma and Ba}{2015}]%
        {kingma2014adam}
\bibfield{author}{\bibinfo{person}{Diederik~P Kingma} {and}
  \bibinfo{person}{Jimmy Ba}.} \bibinfo{year}{2015}\natexlab{}.
\newblock \showarticletitle{Adam: A method for stochastic optimization}. In
  \bibinfo{booktitle}{\emph{International Conference on Learning
  Representations (ICLR)}}.
\newblock


\bibitem[\protect\citeauthoryear{Klementiev, Titov, and Bhattarai}{Klementiev
  et~al\mbox{.}}{2012}]%
        {klementiev2012inducing}
\bibfield{author}{\bibinfo{person}{Alexandre Klementiev}, \bibinfo{person}{Ivan
  Titov}, {and} \bibinfo{person}{Binod Bhattarai}.}
  \bibinfo{year}{2012}\natexlab{}.
\newblock \showarticletitle{Inducing crosslingual distributed representations
  of words}. In \bibinfo{booktitle}{\emph{International Conference ON
  Computational Linguistics (COLING)}}. \bibinfo{pages}{1459--1474}.
\newblock


\bibitem[\protect\citeauthoryear{Meng, Zhang, Xu, and Zhou}{Meng
  et~al\mbox{.}}{2021}]%
        {meng2021learning}
\bibfield{author}{\bibinfo{person}{Qiang Meng}, \bibinfo{person}{Chixiang
  Zhang}, \bibinfo{person}{Xiaoqiang Xu}, {and} \bibinfo{person}{Feng Zhou}.}
  \bibinfo{year}{2021}\natexlab{}.
\newblock \showarticletitle{Learning compatible embeddings}. In
  \bibinfo{booktitle}{\emph{International Conference on Computer Vision
  (ICCV)}}. \bibinfo{pages}{9939--9948}.
\newblock


\bibitem[\protect\citeauthoryear{Mikolov, Le, and Sutskever}{Mikolov
  et~al\mbox{.}}{2013a}]%
        {mikolov2013exploiting}
\bibfield{author}{\bibinfo{person}{Tomas Mikolov}, \bibinfo{person}{Quoc~V Le},
  {and} \bibinfo{person}{Ilya Sutskever}.} \bibinfo{year}{2013}\natexlab{a}.
\newblock \showarticletitle{Exploiting similarities among languages for machine
  translation}.
\newblock \bibinfo{journal}{\emph{arXiv preprint arXiv:1309.4168}}
  (\bibinfo{year}{2013}).
\newblock


\bibitem[\protect\citeauthoryear{Mikolov, Sutskever, Chen, Corrado, and
  Dean}{Mikolov et~al\mbox{.}}{2013b}]%
        {mikolov2013distributed}
\bibfield{author}{\bibinfo{person}{Tomas Mikolov}, \bibinfo{person}{Ilya
  Sutskever}, \bibinfo{person}{Kai Chen}, \bibinfo{person}{Greg~S Corrado},
  {and} \bibinfo{person}{Jeff Dean}.} \bibinfo{year}{2013}\natexlab{b}.
\newblock \showarticletitle{Distributed representations of words and phrases
  and their compositionality}. In \bibinfo{booktitle}{\emph{Advances in Neural
  Information Processing Systems (NeurIPS)}}. \bibinfo{pages}{3111--3119}.
\newblock


\bibitem[\protect\citeauthoryear{Ni, Li, and McAuley}{Ni et~al\mbox{.}}{2019}]%
        {ni2019justifying}
\bibfield{author}{\bibinfo{person}{Jianmo Ni}, \bibinfo{person}{Jiacheng Li},
  {and} \bibinfo{person}{Julian McAuley}.} \bibinfo{year}{2019}\natexlab{}.
\newblock \showarticletitle{Justifying Recommendations using Distantly-Labeled
  Reviews and Fine-Grained Aspects}. In \bibinfo{booktitle}{\emph{Conference on
  Empirical Methods in Natural Language Processing (EMNLP)}}.
  \bibinfo{pages}{188--197}.
\newblock


\bibitem[\protect\citeauthoryear{Rendle, Freudenthaler, Gantner, and
  Schmidt-Thieme}{Rendle et~al\mbox{.}}{2012}]%
        {rendle2012bpr}
\bibfield{author}{\bibinfo{person}{Steffen Rendle}, \bibinfo{person}{Christoph
  Freudenthaler}, \bibinfo{person}{Zeno Gantner}, {and} \bibinfo{person}{Lars
  Schmidt-Thieme}.} \bibinfo{year}{2012}\natexlab{}.
\newblock \showarticletitle{BPR: Bayesian personalized ranking from implicit
  feedback}.
\newblock \bibinfo{journal}{\emph{arXiv preprint arXiv:1205.2618}}
  (\bibinfo{year}{2012}).
\newblock


\bibitem[\protect\citeauthoryear{Ruder, Vuli{\'c}, and S{\o}gaard}{Ruder
  et~al\mbox{.}}{2019}]%
        {ruder2019survey}
\bibfield{author}{\bibinfo{person}{Sebastian Ruder}, \bibinfo{person}{Ivan
  Vuli{\'c}}, {and} \bibinfo{person}{Anders S{\o}gaard}.}
  \bibinfo{year}{2019}\natexlab{}.
\newblock \showarticletitle{A survey of cross-lingual word embedding models}.
\newblock \bibinfo{journal}{\emph{Journal of Artificial Intelligence Research}}
   \bibinfo{volume}{65} (\bibinfo{year}{2019}), \bibinfo{pages}{569--631}.
\newblock


\bibitem[\protect\citeauthoryear{Shen, Xiong, Xia, and Soatto}{Shen
  et~al\mbox{.}}{2020}]%
        {shen2020towards}
\bibfield{author}{\bibinfo{person}{Yantao Shen}, \bibinfo{person}{Yuanjun
  Xiong}, \bibinfo{person}{Wei Xia}, {and} \bibinfo{person}{Stefano Soatto}.}
  \bibinfo{year}{2020}\natexlab{}.
\newblock \showarticletitle{Towards backward-compatible representation
  learning}. In \bibinfo{booktitle}{\emph{IEEE Conference on Computer Vision
  and Pattern Recognition (CVPR)}}. \bibinfo{pages}{6368--6377}.
\newblock


\bibitem[\protect\citeauthoryear{Tagowski, Bielak, and Kajdanowicz}{Tagowski
  et~al\mbox{.}}{2021}]%
        {tagowski2021embedding}
\bibfield{author}{\bibinfo{person}{Kamil Tagowski}, \bibinfo{person}{Piotr
  Bielak}, {and} \bibinfo{person}{Tomasz Kajdanowicz}.}
  \bibinfo{year}{2021}\natexlab{}.
\newblock \showarticletitle{Embedding Alignment Methods in Dynamic Networks}.
  In \bibinfo{booktitle}{\emph{International Conference on Computational
  Science}}. Springer, \bibinfo{pages}{599--613}.
\newblock


\bibitem[\protect\citeauthoryear{Wang, Ying, Li, Rao, Subbian, and
  Leskovec}{Wang et~al\mbox{.}}{2021}]%
        {wang2021bipartite}
\bibfield{author}{\bibinfo{person}{Andrew~Z Wang}, \bibinfo{person}{Rex Ying},
  \bibinfo{person}{Pan Li}, \bibinfo{person}{Nikhil Rao},
  \bibinfo{person}{Karthik Subbian}, {and} \bibinfo{person}{Jure Leskovec}.}
  \bibinfo{year}{2021}\natexlab{}.
\newblock \showarticletitle{Bipartite Dynamic Representations for Abuse
  Detection}. In \bibinfo{booktitle}{\emph{Proceedings of the 27th ACM SIGKDD
  Conference on Knowledge Discovery \& Data Mining}}.
  \bibinfo{pages}{3638--3648}.
\newblock


\bibitem[\protect\citeauthoryear{Wang, He, Wang, Feng, and Chua}{Wang
  et~al\mbox{.}}{2019}]%
        {wang2019neural}
\bibfield{author}{\bibinfo{person}{Xiang Wang}, \bibinfo{person}{Xiangnan He},
  \bibinfo{person}{Meng Wang}, \bibinfo{person}{Fuli Feng}, {and}
  \bibinfo{person}{Tat-Seng Chua}.} \bibinfo{year}{2019}\natexlab{}.
\newblock \showarticletitle{Neural graph collaborative filtering}. In
  \bibinfo{booktitle}{\emph{ACM SIGIR conference on Research and development in
  Information Retrieval (SIGIR)}}. \bibinfo{pages}{165--174}.
\newblock


\bibitem[\protect\citeauthoryear{Xing, Wang, Liu, and Lin}{Xing
  et~al\mbox{.}}{2015}]%
        {xing2015normalized}
\bibfield{author}{\bibinfo{person}{Chao Xing}, \bibinfo{person}{Dong Wang},
  \bibinfo{person}{Chao Liu}, {and} \bibinfo{person}{Yiye Lin}.}
  \bibinfo{year}{2015}\natexlab{}.
\newblock \showarticletitle{Normalized word embedding and orthogonal transform
  for bilingual word translation}. In \bibinfo{booktitle}{\emph{North American
  Chapter of the Association for Computational Linguistics (NAACL)}}.
  \bibinfo{pages}{1006--1011}.
\newblock


\bibitem[\protect\citeauthoryear{Yao, Sun, Ding, Rao, and Xiong}{Yao
  et~al\mbox{.}}{2018}]%
        {yao2018dynamic}
\bibfield{author}{\bibinfo{person}{Zijun Yao}, \bibinfo{person}{Yifan Sun},
  \bibinfo{person}{Weicong Ding}, \bibinfo{person}{Nikhil Rao}, {and}
  \bibinfo{person}{Hui Xiong}.} \bibinfo{year}{2018}\natexlab{}.
\newblock \showarticletitle{Dynamic word embeddings for evolving semantic
  discovery}. In \bibinfo{booktitle}{\emph{Proceedings of the eleventh acm
  international conference on web search and data mining}}.
  \bibinfo{pages}{673--681}.
\newblock


\bibitem[\protect\citeauthoryear{Ying, He, Chen, Eksombatchai, Hamilton, and
  Leskovec}{Ying et~al\mbox{.}}{2018}]%
        {ying2018graph}
\bibfield{author}{\bibinfo{person}{Rex Ying}, \bibinfo{person}{Ruining He},
  \bibinfo{person}{Kaifeng Chen}, \bibinfo{person}{Pong Eksombatchai},
  \bibinfo{person}{William~L Hamilton}, {and} \bibinfo{person}{Jure Leskovec}.}
  \bibinfo{year}{2018}\natexlab{}.
\newblock \showarticletitle{Graph Convolutional Neural Networks for Web-Scale
  Recommender Systems}. In \bibinfo{booktitle}{\emph{ACM SIGKDD Conference on
  Knowledge Discovery and Data Mining (KDD)}}. \bibinfo{pages}{974--983}.
\newblock


\bibitem[\protect\citeauthoryear{Zhang, Xu, Kawarabayashi, Jegelka, and
  Boyd-Graber}{Zhang et~al\mbox{.}}{2019}]%
        {zhang2019girls}
\bibfield{author}{\bibinfo{person}{Mozhi Zhang}, \bibinfo{person}{Keyulu Xu},
  \bibinfo{person}{Ken-ichi Kawarabayashi}, \bibinfo{person}{Stefanie Jegelka},
  {and} \bibinfo{person}{Jordan Boyd-Graber}.} \bibinfo{year}{2019}\natexlab{}.
\newblock \showarticletitle{Are Girls Neko or Sh$\backslash$= ojo?
  Cross-Lingual Alignment of Non-Isomorphic Embeddings with Iterative
  Normalization}.
\newblock \bibinfo{journal}{\emph{arXiv preprint arXiv:1906.01622}}
  (\bibinfo{year}{2019}).
\newblock


\bibitem[\protect\citeauthoryear{Zhu, Zhao, Yang, Lin, Zhou, Ai, Li, and
  Zhou}{Zhu et~al\mbox{.}}{2019}]%
        {zhu2019aligraph}
\bibfield{author}{\bibinfo{person}{Rong Zhu}, \bibinfo{person}{Kun Zhao},
  \bibinfo{person}{Hongxia Yang}, \bibinfo{person}{Wei Lin},
  \bibinfo{person}{Chang Zhou}, \bibinfo{person}{Baole Ai},
  \bibinfo{person}{Yong Li}, {and} \bibinfo{person}{Jingren Zhou}.}
  \bibinfo{year}{2019}\natexlab{}.
\newblock \showarticletitle{Aligraph: A comprehensive graph neural network
  platform}.
\newblock \bibinfo{journal}{\emph{arXiv preprint arXiv:1902.08730}}
  (\bibinfo{year}{2019}).
\newblock


\end{thebibliography}

\newpage
\appendix
\newpage
\section{Additional Implementation Details}

\subsection{Embedding Models}
\label{app:emb_model}
We evolve the GraphSAGE embedding models as follows.
For the Musical Instruments and Video Games, we start with a 2-layer GraphSAGE model with the hidden dimensionality of 256 at $t_0$. Then, we increase the layer size to 3 at $t_2$ and increase the hidden dimensionality by 64 at every timestamp.
For the Grocery dataset, we use smaller models due to the limited GPU memory; we start with a 1-layer GraphSAGE model with hidden diemensionality of 256 at $t_0$. Then, we increase the layer size to 2 at $t_2$ and increase the hidden dimensionality by 64 until $t_2$.
All the models are trained for 500 epochs with Adam~\citep{kingma2014adam}, with a learning rate of 0.001, and the weight decay of 0.01.

\subsection{Unintended Tasks}
\label{app:detail_unintended}

In designing unintended tasks, we utilize the review rating information associated with each edge, which takes an integer value of between 1 and 5. Below we explain each unintended task, how training is performed on ver-0 embeddings and how predictions are made at each timestamp. 
We let $V_k^{\rm (user)}$ and  $V_k^{\rm (item)}$ denote the set of users and items appearing at least once in $E_k$.

\begin{itemize}
    \item \textbf{User activity prediction:} Given a user embedding obtained at $t_k$, we predict whether the user will interact with at least a single item between $t_k$ and $t_{k+1}$. At training time, we train on users in $V_0^{\rm (user)}$ at $t_0$, and validate on users in $V_1^{\rm (user)}$ at $t_1$.
    At test time $t_k, 2 \leq k$, we make predictions on users in $V_k^{\rm (user)}$. 
    
    \item \textbf{User positive activity prediction:} The task is the same as the above, except that we predict whether a user will have at least one positive interaction with an item (\ie, rating $\geq 4$) or not.
    
    \item \textbf{Item rating avg prediction:} Given an item embedding obtained at $t_k$, predict the average rating of the item until $t_{k+1}$, where we only consider items that receive more than 10 reviews until $t_k$. 
    We binarize the average rating by thresholding it at the median value at $t_0$.
    At training time $t_0$, we train on items in $V_0^{\rm (item)}$ for the average item rating until $t_0$ and validate on the average item rating until $t_1$.
    At test time $t_k, 1 \leq k$, we make predictions on items in  $V_k^{\rm (item)}$.
    
    \item \textbf{Item rating std prediction:} The task is the same as the above, except that we predict the standard deviation of the item ratings. The standard deviation is binarized by thresholding at 1.
    
    \item \textbf{Edge rating prediction:} Given a pair of user and item embeddings at $t_k$, predict whether the user gives a positive rating (\ie, $\ge 4$) to the item between $t_k$ and $t_{k+1}$.
    At training time $t_0$, we train on edges in $E_0$ and validate on edges in $E_1 \setminus E_0$.
    At test time $t_k, 1 \leq k \leq K$, we make predictions on edges in $E_{k+1} \setminus E_k$.
\end{itemize}

Note that the test prediction for the first two tasks is performed for $t_k, 2 \leq k$, while the test prediction for the last three tasks is performed for $t_k, 1 \leq k$. This is because the first two tasks are predicting future activity of existing users.

All the tasks are binary classification, and we use ROC-AUC on the test set series as the performance metric.
All consumer models are 1-hidden-layer MLP models and are trained on ver-0 embeddings generated by $M_0$. For each task, we tune the hidden embedding dimensionality from $\{128, 256, 512, 1024\}$, the dropout ratio from $\{0, 0.25, 0.5\}$, and performed early-stopping based on performance on the validation set.
We report the unintended task performance averaged over 10 random seeds.

\end{document}